\documentclass[letterpaper, 10 pt, conference]{ieeeconf}  % Comment this line out if you need a4paper

\IEEEoverridecommandlockouts                              % This command is only needed if 
\usepackage{graphics} % for pdf, bitmapped graphics files
\usepackage{epsfig} % for postscript graphics files
\usepackage{times} % assumes new font selection scheme installed
\usepackage{amsmath,amssymb,amsfonts}
\usepackage{multirow}
\usepackage{tabularx}
\usepackage{balance}

\usepackage{calc}

\usepackage{caption}
\DeclareCaptionLabelSeparator{dottilde}{.~~}
\usepackage{hyperref}
\usepackage{cleveref}
\usepackage{booktabs}

\usepackage{algorithm}
\usepackage{algpseudocode}
\usepackage{amsmath}

\usepackage{pifont}

\usepackage{booktabs}
\usepackage{makecell}  
\usepackage{graphicx}  
\usepackage{color} % \textcolor
\usepackage{soul} % \st

\usepackage{booktabs}
\usepackage{tabularx}
\usepackage{array}
\usepackage{pifont}

\newcommand{\benchyes}{\ding{51}}
\newcommand{\benchno}{\ding{55}}

\usepackage{amssymb}

\title{\LARGE \bf
VP-EVT: Active Visibility Preservation for Embodied Visual Tracking
}

\author{
Yanci Wen$^{1}$,
Tianle Zeng$^{1}$,
Jianwei Peng$^{2}$,
Weixi Situ$^{1}$,
Hanjing Ye$^{1,3}$,
and Hong Zhang$^{1,\dagger}$
\thanks{$^{1}$Shenzhen Key Laboratory of Robotics and Computer Vision, Southern University of Science and Technology, Shenzhen, China.}
\thanks{$^{2}$Southern University of Science and Technology, Shenzhen, China.}
\thanks{$^{3}$Tencent Robotics X, Shenzhen, China.}
\thanks{$^{\dagger}$Corresponding author: Hong Zhang (hzhang@sustech.edu.cn).}
}

\begin{document}

\maketitle
\thispagestyle{empty}
\pagestyle{empty}

\begin{abstract}
    Embodied Visual Tracking (EVT) is an essential capability for intelligent robots in human--robot collaboration, requiring an agent to follow a designated target person in dynamic environments relying solely on egocentric visual perception. However, persistent pedestrian occlusions in dynamic scenes pose critical challenges to existing EVT methods. Lacking proactive visibility preservation, reactive tracking methods frequently experience prolonged target loss under severe occlusions. Therefore, we propose \textbf{VP-EVT}, a framework that treats target visibility as an explicit objective of motion control rather than a by-product of following. Acting on this objective requires understanding both what obstructs the target and how robot motion can change that obstruction. VP-EVT therefore uses language-guided perception to extract separate target and pedestrian observations. When occlusion makes current observations incomplete, VP-EVT combines visual history with ego-motion to maintain estimates of their relative positions and motion. Using these estimates and the current following waypoints, VP-EVT predicts whether continued following will sustain the obstruction. When the predicted risk is low, the original following motion is retained; otherwise, a recurrent controller combines updated estimates with maneuver context from preceding decisions to coordinate viewpoint adjustment and the return to stable following. 
    Moreover, we introduce \textbf{OccEVT-Bench} for closed-loop evaluation of persistent occlusion and collect expert demonstrations for training. On its Risk split, VP-EVT improves success rate by 121.4\% relative to a competitive baseline and reduces visibility deficit and longest target-loss duration by 78.8\% and 74.4\%, respectively.
\end{abstract}
\section{INTRODUCTION}

\begin{figure}[t]
    \centering
    \includegraphics[width=\linewidth]{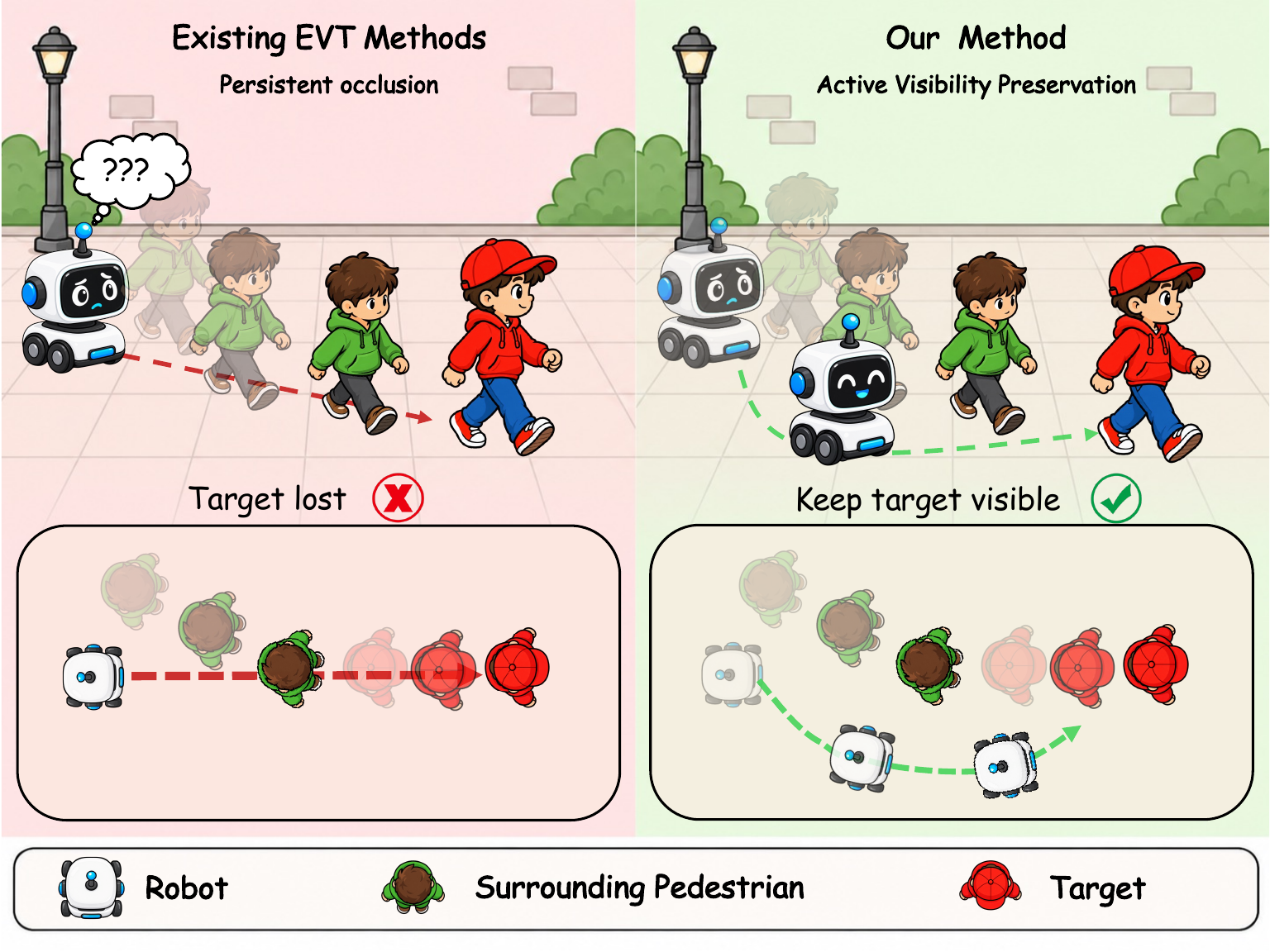}
    \caption{
        Standard following and \textbf{VP-EVT} under dynamic
        pedestrian occlusion.
        Continuing behind an occluding pedestrian can prolong
        target loss, whereas visibility-preserving motion
        changes the robot's viewpoint to reduce the obstruction
        while continuing to follow the target.
        The upper and lower panels provide perspective and
        bird's-eye illustrations, respectively.
    }
    \label{fig:figure1}
    \vspace*{-0.25in}
\end{figure}

Embodied Visual Tracking (EVT) requires an agent to follow a designated moving target using egocentric visual observations \cite{luo2018active,zhong2021distraction,zhong2024evt,wang2025trackvla}. However, when target visibility is treated as a by-product of following rather than an explicit motion objective, the agent may continue following without addressing an obstructed view. As illustrated in Fig.~\ref{fig:figure1}, a robot in a dynamic pedestrian scene can maintain the desired following distance and avoid collisions while remaining behind a pedestrian that blocks the target. Continuing along the same following motion can sustain this obstruction, leading to prolonged target loss and eventual tracking failure even when a viewpoint change could improve visibility. The challenge is therefore to anticipate when following motion will prolong pedestrian occlusion and actively adjust the viewpoint while continuing to follow the target.

Robust target estimation and effective recovery do not by themselves prevent persistent physical occlusion. Recognition, localization, and temporal memory preserve target information under degraded observations \cite{maalouf2024followanything,ye2023robotpersonfollowing,ye2024personreidentification,liu2025trackvlapp}, but even an accurate target estimate does not specify how the robot should move to avoid remaining behind an occluding pedestrian. Recovery mechanisms intervene when tracking becomes unreliable \cite{ye2025rpfsearch,sun2026oavat,wu2025vlmassistant}, responding to degraded visibility rather than explicitly predicting whether continued following will prolong the obstruction. The missing decision is therefore whether the current following motion will sustain the blockage and should be changed. These limitations motivate treating target visibility as an explicit objective of motion control, so that the robot can adjust its following motion to prevent avoidable occlusion from persisting.

We propose \textbf{VP-EVT}, an active visibility-preserving framework. To understand which pedestrian may obstruct the target, VP-EVT first uses language-conditioned grounding to extract their visual evidence separately. As occlusion makes this evidence incomplete, we integrate successive observations with ego-motion into target--pedestrian temporal beliefs, maintaining uncertainty-aware spatial and motion estimates even when direct observations are unavailable. Because persistent occlusion also depends on the agent's future motion, VP-EVT evaluates these beliefs along the current following waypoints to determine whether intervention is needed. VP-EVT selectively adjusts its motion to preserve target visibility and restore stable following. Evaluating this capability requires distinguishing reduced persistent occlusion from eventual target recovery. Aggregate success and tracking scores~\cite{wang2025trackvla} alone cannot distinguish these outcomes or reveal whether reduced physical occlusion comes at the cost of out-of-view target loss. 

To bridge this evaluation gap, we introduce \textbf{OccEVT-Bench}, a dedicated closed-loop benchmark for occlusion-aware embodied visual tracking that comprehensively quantifies visibility loss alongside tracking safety. Complemented by this platform, we contribute \textbf{OccEVT-Expert}, an expert demonstration dataset containing 20,000 dynamic-occlusion episodes for training proactive visibility-preserving policies.
\section{RELATED WORK}

\subsection{Embodied Visual Tracking}

Embodied visual tracking couples target perception with motion control through egocentric observations. Learning-based approaches learn tracking control directly from visual inputs \cite{luo2018active,luo2020endtoend,dionigi2024dvat}, while adversarial interactions improve robustness to challenging target motion, obstacles, and distractors \cite{zhong2019advat,zhong2021advatplus,zhong2021distraction}. Visual foundation models provide transferable target representations and flexible target specification, supporting generalization through offline policy learning \cite{maalouf2024followanything,zhong2024evt}. Vision--language--action models further integrate language-conditioned perception with action generation \cite{zhang2025uninavid,wang2025trackvla}. Competitive multi-agent training and context-conditioned adaptation extend these policies toward adaptive interference and cross-embodiment tracking, respectively \cite{liu2026comatrack,wu2026adatracker}.

Under occlusion, observation-centered approaches exploit visible body regions, temporal memory, and confidence-aware filtering to maintain target estimates \cite{ye2023robotpersonfollowing,liu2025trackvlapp,sun2026oavat}. Although these mechanisms retain information when observations become incomplete, they do not by themselves remove a physical obstruction. Search- and recovery-oriented approaches couple tracking with exploration or generate recovery motion through belief-guided search, learned trajectory planning, and vision--language reasoning \cite{dionigi2022evat,ye2025rpfsearch,sun2026oavat,wu2025vlmassistant}. These approaches emphasize acquiring or regaining reliable target observations.

Predictive control considers how future motion affects tracking. Belief-space policy learning addresses target-state uncertainty under limited fields of view \cite{yang2023policy}, while structured approaches combine scene reconstruction with target-trajectory prediction \cite{zhong2023rspt}. Geometry-based planners additionally incorporate visibility and occlusion constraints into tracking trajectories \cite{ji2022elastic,situ2025adaprpf}. Learned world models offer another route by evaluating candidate actions through future target-evidence rollouts \cite{hu2026cstwm}. Our framework adopts a complementary relational perspective, using persistent target beliefs and short-term pedestrian estimates to explicitly represent the interaction underlying occlusion. Conditioned on the current following waypoints, it anticipates whether the obstruction will persist and coordinates selective visibility-preserving motion with a return to stable following.

\begin{figure*}[!t]
    \centering
    \includegraphics[width=\textwidth]{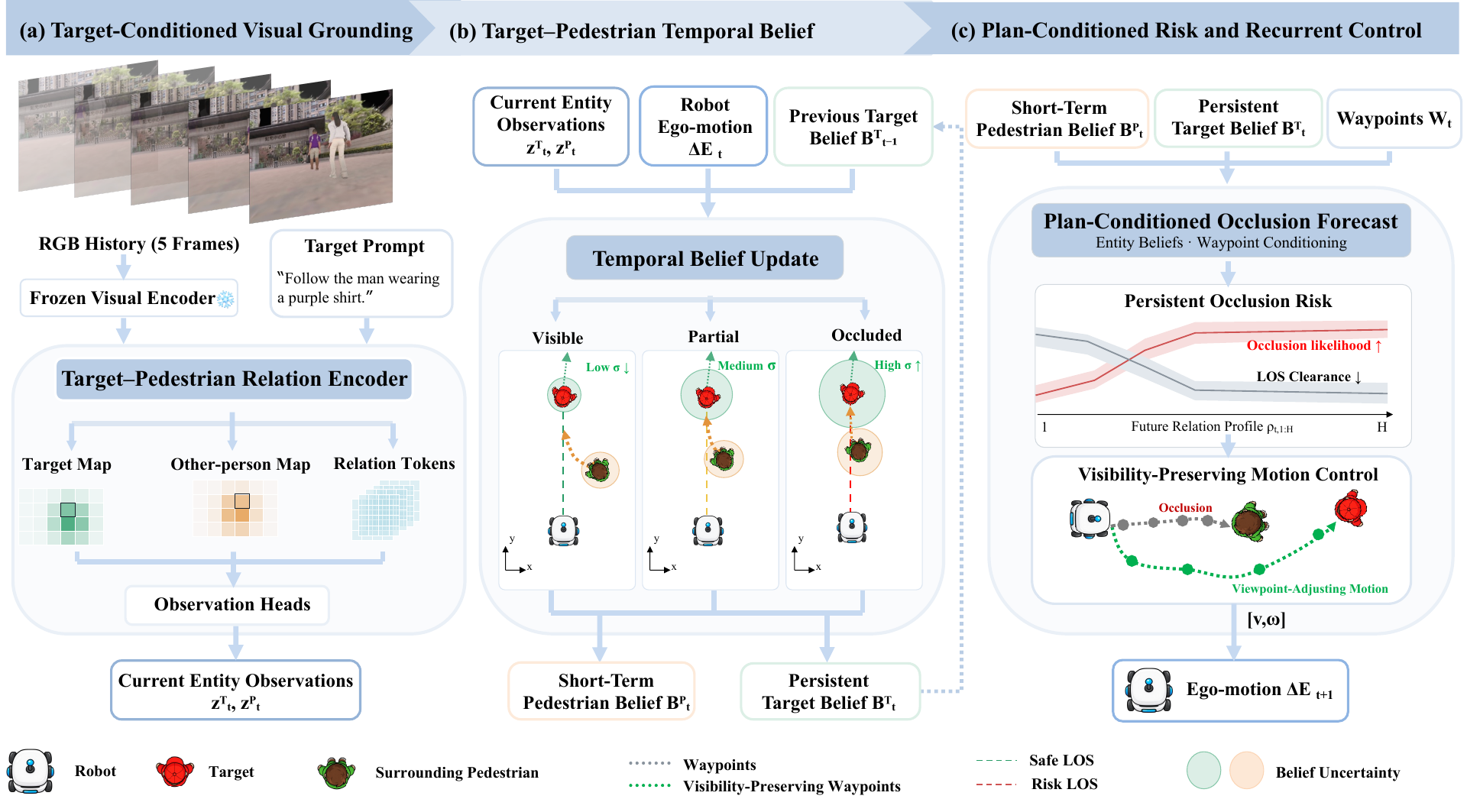}
    \caption{
        \textbf{Overview of VP-EVT.}
        (a) A causal RGB history and a language prompt produce
        \emph{Current Entity Observations}:
        $\mathbf{z}_t^T$ for the designated target ($T$)
        and $\mathbf{z}_t^P$ for a potential occluding
        pedestrian ($P$).
        (b) Visual observations and robot ego-motion
        $\Delta\mathbf{E}_t$ update a persistent target belief
        $\mathbf{B}_t^T$ and a short-term pedestrian belief
        $\mathbf{B}_t^P$.
        Uncertainty increases when visual evidence becomes
        unavailable.
        (c) The target and pedestrian beliefs, together with
        following waypoints $\mathbf{W}_t$, condition
        persistent-occlusion risk prediction and recurrent
        visibility-preserving control.
        The curves schematically illustrate visibility
        degradation, while the motion example contrasts
        continued following with a visibility-preserving
        maneuver.
        The controller produces linear and angular velocity
        commands $[v,\omega]$, and the resulting ego-motion
        $\Delta\mathbf{E}_{t+1}$ informs the next belief update.
    }
    \label{fig:relational_belief}
\end{figure*}

\subsection{Visual Tracking Benchmarks}

Pre-recorded datasets evaluate tracking across diverse targets and long video sequences \cite{fan2019lasot,huang2021got10k}, with specialized datasets examining heavy occlusion, containment, and target disappearance \cite{sahin2023hoot,vanhoorick2023tcow,ye2025tptbench}. These evaluations characterize perception robustness but do not allow the evaluated agent to alter subsequent viewpoints. Closed-loop simulation platforms support interactive tracking across diverse scenes and embodiments \cite{zhong2025unrealzoo}. Existing protocols cover dynamic distractors and adversarial interference \cite{zhong2021distraction,liu2026comatrack}, as well as visibility, safety, and comfort in crowd-aware following \cite{ye2025followbench}.

A representative language-conditioned benchmark reports success, tracking, and collision rates under pedestrian interference \cite{wang2025trackvla}. These aggregate metrics capture overall performance but do not explicitly characterize target-loss persistence or distinguish physical occlusion from out-of-view failures. OccEVT-Bench complements these evaluations with controlled pedestrian interactions that can produce persistent occlusion under standard following. It measures target-loss persistence and accumulated visibility degradation alongside stable following recovery and collision safety, reporting physical-occlusion and out-of-view losses separately. Matched Safe episodes additionally assess whether occlusion-handling behavior disrupts otherwise valid following. Benchmark construction and metrics are described in Sec.~\ref{sec:benchmark}.
\section{METHODOLOGY}
\label{sec:method}

We study visibility-preserving EVT under pedestrian occlusion. The objective is to minimize cumulative visibility deficit while avoiding collisions and maintaining or restoring the prescribed following distance and target-relative pose.

As illustrated in Fig.~\ref{fig:relational_belief}, VP-EVT comprises three complementary modules: (a) \emph{Target-Conditioned Visual Grounding}, (b) \emph{Target--Pedestrian Temporal Belief}, and (c) \emph{Plan-Conditioned Risk and Recurrent Control}.

At decision time $t$, the system receives a causal RGB history $\mathcal{I}_t$, a target description $q$, and robot ego-motion $\Delta\mathbf{E}_t$. A frozen follower~\cite{omtrackvla2025} provides robot-local following waypoints $\mathbf{W}_t=\{\mathbf{w}_{t,k}\}_{k=1}^{H}$, where $H$ denotes the number of waypoints.

First, visual grounding uses a frozen encoder and prompt-guided target and other-person maps to produce observations $\mathbf{z}_t^T$ and $\mathbf{z}_t^P$ (Fig.~\ref{fig:relational_belief}(a)). Next, these observations, ego-motion, and prior estimates update a persistent target belief $\mathbf{B}_t^T$ and a short-term pedestrian belief $\mathbf{B}_t^P$, maintaining uncertainty-aware position and motion estimates under incomplete observations (Fig.~\ref{fig:relational_belief}(b)). Finally, the beliefs are evaluated along the following waypoints to anticipate persistent occlusion (Fig.~\ref{fig:relational_belief}(c)). Standard following is retained when intervention is unnecessary; otherwise, a recurrent controller generates linear and angular velocity commands $[v,\omega]$ to adjust the viewpoint until stable following resumes. Updated observations and ego-motion feed back into belief estimation, closing the loop.

\subsection{Target-Conditioned Visual Grounding}
\label{sec:visual_grounding}

As shown in Fig.~\ref{fig:relational_belief}(a),
a frozen visual encoder extracts spatial features
from a causal five-frame RGB history.
A target--pedestrian relation encoder uses the
language prompt to produce a target map and an
other-person map, retaining visual evidence about
both the designated target and potential occluders.
The maps are not constrained to be complementary,
since a coarse image region may contain both
the target and an occluding pedestrian during
partial occlusion.

These image-space maps weight the relation tokens
to obtain separate features for the target and
pedestrian.
For each map, we additionally retain its centroid,
spatial spread, and confidence statistics.
Temporal observation heads combine these features
over the RGB history with robot-motion information
to estimate the target and pedestrian locations
relative to the robot.

Specifically, each observation head predicts
a joint distribution over discretized bearing
and distance:
\begin{equation}
    P_t^e(i,j)
    =
    p(\theta_i,d_j
    \mid \mathcal{I}_t,q,\mathbf{M}_t),
    \quad e\in\{T,P\},
    \label{eq:polar_observation}
\end{equation}
where $T$ and $P$ index the target and pedestrian,
and $\theta_i$ and $d_j$ denote bearing and
distance bins, respectively.
This distribution is learned from robot-relative
position supervision; it is not obtained by
geometrically projecting the image heatmaps
using depth measurements.

The bearing--distance hypotheses are converted
into local Cartesian positions using
$[x,y]=[d\cos\theta,d\sin\theta]$, with learned
within-bin refinements.
Their probability-weighted moments provide
a position estimate and its covariance.
Together with predicted planar velocity and
visibility or observation availability, they form
the current target and pedestrian observations
$\mathbf{z}_t^T$ and $\mathbf{z}_t^P$.

These observations connect image-space grounding
to robot-local spatial estimation.
The temporal belief module subsequently filters
them and encodes the resulting distributions
as coarse robot-centric belief maps.

\subsection{Target--Pedestrian Temporal Belief}
\label{sec:temporal_belief}

Given the current observations $\mathbf{z}_t^T$ and
$\mathbf{z}_t^P$, we maintain robot-local temporal beliefs
that retain spatial estimates when visual evidence
becomes unavailable.
As illustrated in Fig.~\ref{fig:relational_belief}(b),
the module maintains a persistent target belief and
an observation-driven, short-term pedestrian belief
to capture the evolving occlusion interaction.

\textbf{Temporal state estimation.}
For the target and pedestrian, indexed by
$e\in\{T,P\}$, we represent position and velocity
with a Gaussian belief:
\begin{equation}
    \mathbf{x}_t^e
    = [x_t^e,y_t^e,v_{x,t}^e,v_{y,t}^e]^\top,
    \quad
    p(\mathbf{x}_t^e)
    = \mathcal{N}(\boldsymbol{\mu}_t^e,\boldsymbol{\Sigma}_t^e),
    \label{eq:state_belief}
\end{equation}
where $(x_t^e,y_t^e)$ and $(v_{x,t}^e,v_{y,t}^e)$
denote planar position and velocity expressed
in the robot-local coordinate frame, respectively.
The mean $\boldsymbol{\mu}_t^e$ is the estimated
state, and the covariance
$\boldsymbol{\Sigma}_t^e$ captures its uncertainty.

The target belief combines motion propagation
with visual correction.
The previous estimate is first propagated using
its velocity and transformed into the current
robot frame using ego-motion
$\Delta\mathbf{E}_t$.
Reliable visual observations then correct this
prediction.
When target evidence is unavailable, the belief
continues to propagate with increasing uncertainty
until observations resume.
The visible, partial, and occluded examples in
Fig.~\ref{fig:relational_belief}(b) illustrate
how reduced visual evidence affects the spatial
estimate and its uncertainty.

The pedestrian belief prioritizes available
observations to capture its current position
and motion, while temporal propagation bridges
temporary observation gaps.
This preserves target continuity through occlusion
without unnecessarily smoothing the motion of
the pedestrian causing it.

\textbf{Coarse spatial belief encoding.}
We convert each filtered position distribution
into a coarse robot-centric belief map on a
polar grid discretized by bearing and distance.
Let $\mathbf{c}_m$ denote the Cartesian position
of grid-cell center $m$, and let
$\boldsymbol{\mu}_{p,t}^e$ and
$\boldsymbol{\Sigma}_{p,t}^e$ denote the position
components of $\boldsymbol{\mu}_t^e$ and
$\boldsymbol{\Sigma}_t^e$, respectively.
For
$\boldsymbol{\delta}_{t,m}^e
=\mathbf{c}_m-\boldsymbol{\mu}_{p,t}^e$,
we compute spatial weights from Gaussian scores
at the grid centers:
\begin{equation}
    b_t^e(m) \propto
    \exp\!\left[
    -\frac{1}{2}
    (\boldsymbol{\delta}_{t,m}^e)^\top
    (\boldsymbol{\Sigma}_{p,t}^e)^{-1}
    \boldsymbol{\delta}_{t,m}^e
    \right].
    \label{eq:coarse_belief_map}
\end{equation}
Here, $b_t^e(m)$ denotes the spatial weight at
grid-cell center $m$, normalized such that
$\sum_m b_t^e(m)=1$.
These weights encode plausible target or pedestrian
locations, providing a coarse spatial representation
for subsequent risk prediction and control.
A confident estimate concentrates the spatial
weights, while greater uncertainty spreads
them across multiple cells.

Finally, each coarse map is combined with
the corresponding position, velocity, uncertainty,
and visibility or observation availability
to form belief tokens
$\mathbf{B}_t^T$ and $\mathbf{B}_t^P$.
These tokens provide the spatial representation
used by subsequent risk prediction and
visibility-preserving control.

\subsection{Plan-Conditioned Risk and Recurrent Control}
\label{sec:risk_control}

Given the target and pedestrian beliefs, this
module evaluates the following waypoints and
selectively adjusts robot motion to preserve
visibility, as illustrated in
Fig.~\ref{fig:relational_belief}(c).

\textbf{Plan-conditioned occlusion risk.}
Whether a pedestrian causes persistent occlusion
depends on the relative positions and motions
of the robot, target, and pedestrian.
We therefore predict risk from the two belief
tokens, the current following waypoints, and
robot-motion history:
\begin{equation}
    r_t =
    f_r(\mathbf{B}_t^T,\mathbf{B}_t^P,
        \mathbf{W}_t,\mathbf{M}_t),
    \label{eq:occlusion_risk}
\end{equation}
where $f_r$ is a learned risk predictor,
$\mathbf{M}_t$ denotes recent robot-motion history,
and $r_t\in[0,1]$ represents the predicted risk
of persistent physical target occlusion.
Their joint representation allows risk prediction
to account for where the target and pedestrian
are likely to be and how the robot intends
to move.

The classifier is supervised by the duration
of subsequent physical target loss in
standard-following rollouts.
Positive labels identify sustained loss while
the target remains inside the camera field of
view, distinguishing persistent obstruction
from brief overlap or out-of-view failure.
When predicted risk is low, standard following
continues. Temporally confirmed high risk
activates visibility-preserving control.

\textbf{Recurrent visibility-preserving control.}
An intervention must move the robot away from
an obstructed viewpoint and subsequently restore
a suitable following position.
We generate this motion through short control
chunks, conditioned on updated target and
pedestrian beliefs and a recurrent state that
connects successive decisions.
The controller encodes the current inputs,
updates its memory, and predicts control
increments:
\begin{equation}
\begin{aligned}
    \mathbf{o}_t
        &= \mathcal{E}_{\theta}\!\left(
           \mathbf{B}_t^T,\mathbf{B}_t^P,
           \mathbf{W}_t,\mathbf{u}_t^{\mathrm{cur}}
           \right),\\
    \mathbf{h}_t
        &= \mathcal{R}_{\theta}\!\left(
           \mathbf{h}_{t-1},\mathbf{o}_t
           \right),\\
    \Delta\mathbf{U}_t
        &= \mathcal{D}_{\theta}\!\left(
           \mathbf{h}_t,\bar{\mathbf{W}}_t,
           \bar{\mathbf{G}}_t
           \right).
\end{aligned}
\label{eq:recurrent_control}
\end{equation}
Here, $\mathcal{E}_{\theta}$ encodes the current
inputs and the relative positions and velocities
derived from the two beliefs into a control
context $\mathbf{o}_t$. $\mathbf{u}_t^{\mathrm{cur}}$ specifies
the current linear and angular velocity commands.
The recurrent update $\mathcal{R}_{\theta}$,
implemented with a GRU, combines the previous
state with the current context to obtain
$\mathbf{h}_t$.
The decoder $\mathcal{D}_{\theta}$ then predicts
the next control increments.
Together, $\theta$ denotes the controller's
learnable parameters.

To condition individual controls within a chunk,
we resample the following waypoints into
references $\bar{\mathbf{W}}_t$ and extrapolate
the estimated target and pedestrian positions
over the execution interval.
Their predicted positions, relative displacement,
and velocities form $\bar{\mathbf{G}}_t$.
These references guide control generation using
both the proposed following direction and
the predicted target--pedestrian interaction.

The normalized increments
$\Delta\mathbf{U}_t$ are scaled and accumulated
from $\mathbf{u}_t^{\mathrm{cur}}$ to produce
$\mathbf{U}_t=\{[v_{t,j},\omega_{t,j}]\}_{j=1}^{K}$,
subject to velocity and acceleration limits.
The controller predicts $K=4$ controls and
executes the complete chunk before the next
decision.

The recurrent state is initialized when
intervention begins and retained across
successive chunks.
Each new chunk incorporates updated visual
beliefs while remaining conditioned on the
preceding maneuver.
The green path in
Fig.~\ref{fig:relational_belief}(c) illustrates
the resulting closed-loop motion.

\textbf{Restoring standard following.}
Restoring visibility alone does not ensure that the robot has regained a suitable following position. A separate handoff predictor therefore uses the target and pedestrian beliefs and their predicted relations along the following waypoints to determine when standard following can resume. The robot returns to waypoint following once the handoff criterion and the visibility and following-geometry checks are satisfied across consecutive decisions.

\paragraph{Training}
We train VP-EVT in stages on OccEVT-Expert, keeping the pretrained visual backbone and base follower frozen. Target and pedestrian maps, states, and visibility annotations supervise visual grounding and temporal belief estimation. Observation heads are supervised only when visual evidence is available, whereas belief supervision extends through occlusion. The risk and handoff predictors are trained separately using class-weighted binary cross-entropy.

With the perception and decision modules fixed, we train the recurrent controller on complete expert maneuver sequences, retaining its hidden state across consecutive decisions. The objective combines motion imitation with human-clearance and following-distance penalties:
\begin{equation}
\mathcal{L}_{\mathrm{ctrl}}
= \mathcal{L}_{\mathrm{imit}}
+ \lambda_c \mathcal{L}_{\mathrm{clear}}
+ \lambda_f \mathcal{L}_{\mathrm{follow}},
\label{eq:control_loss}
\end{equation}
where $\mathcal{L}_{\mathrm{imit}}$ applies Smooth-L1 losses to expert control increments, velocity commands, and the resulting robot trajectories. $\mathcal{L}_{\mathrm{clear}}$ penalizes insufficient human clearance, while $\mathcal{L}_{\mathrm{follow}}$ penalizes deviations from the desired following distance; $\lambda_c$ and $\lambda_f$ balance these terms. Privileged annotations are used only for supervision, not as policy inputs at inference.

\section{BENCHMARK DESIGN}
\label{sec:benchmark}

\begin{figure}[t]
\centering
\includegraphics[width=\linewidth]{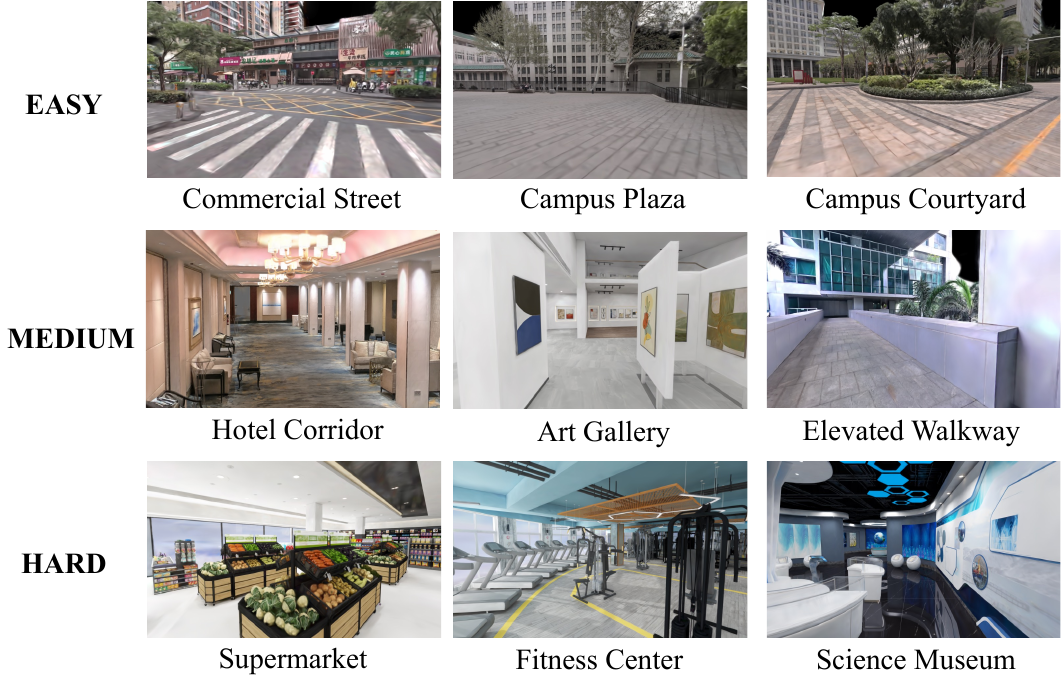}
\caption{\textbf{Overview of OccEVT-Bench.}
Episodes span indoor and outdoor scenes, diverse target
appearances and motions, and dynamic pedestrian interactions.
Risk episodes assess persistent-occlusion handling,
while complementary Safe episodes assess continued following
without unnecessary intervention.
Easy, Medium, and Hard levels cover different interaction
difficulties.}
\label{fig:benchmark}
\end{figure}

To distinguish active visibility preservation from eventual
target recovery, we constructed \textbf{OccEVT-Bench}, a
closed-loop benchmark centered on persistent pedestrian
occlusion. It evaluates whether an embodied tracker can
reduce sustained obstruction through its own motion while
maintaining effective following.
Built on Habitat-GS~\cite{xia2026habitatgs}, the platform
supports 3D Gaussian Splatting scene rendering and
custom-scene import.
Table~\ref{tab:benchmark_comparison} summarizes its scope
relative to existing benchmarks.

\subsection{Episode Design}

Risk episodes are designed to produce persistent
pedestrian occlusion under reference following, whereas
complementary Safe episodes contain interactions
without such occlusion.
They assess occlusion mitigation and disruption to
otherwise valid following, respectively.
Episode labels remain fixed across evaluated policies.

OccEVT-Bench contains 400 evaluation episodes
spanning diverse scenes and pedestrian interactions.
As illustrated in Fig.~\ref{fig:benchmark}, episodes
vary in target appearance, motion patterns,
and interaction difficulty.

\begin{table}[t]
\centering
\caption{\textbf{Comparison of tracking benchmarks.}}
\label{tab:benchmark_comparison}

\footnotesize
\setlength{\tabcolsep}{2.5pt}
\renewcommand{\arraystretch}{1.12}

\begin{tabularx}{\columnwidth}{
@{}>{\raggedright\arraybackslash}X *{6}{c}@{}}
\toprule
\textbf{Benchmark}
& \textbf{RGB-CL}
& \textbf{Occ.}
& \textbf{LTL}
& \textbf{Attr.}
& \textbf{Safety}
& \textbf{Scene} \\
\midrule

HOOT~\cite{sahin2023hoot}
& \benchno
& \benchyes
& \benchno
& \benchno
& \benchno
& Video \\

TCOW~\cite{vanhoorick2023tcow}
& \benchno
& \benchyes
& \benchno
& \benchno
& \benchno
& Video \\

TPT-Bench~\cite{ye2025tptbench}
& \benchno
& \benchyes
& \benchno
& \benchno
& \benchno
& Video \\

\midrule

EVT-Bench~\cite{wang2025trackvla}
& \benchyes
& \benchno
& \benchno
& \benchno
& \benchyes
& Mesh \\

Follow-Bench~\cite{ye2025followbench}
& \benchno
& \benchyes
& \benchno
& \benchno
& \benchyes
& 2D \\

\midrule

\textbf{OccEVT-Bench}
& \benchyes
& \benchyes
& \benchyes
& \benchyes
& \benchyes
& 3DGS \\

\bottomrule
\end{tabularx}

\vspace{2pt}
\parbox{\columnwidth}{\scriptsize
\textbf{RGB-CL}: closed-loop evaluation with RGB inputs;
\textbf{Occ.}: occlusion-focused evaluation or explicit
visibility metrics;
\textbf{LTL}: longest consecutive target-loss duration
reported as a performance metric;
\textbf{Attr.}: physical-occlusion and out-of-FOV loss
durations reported separately;
\textbf{Safety}: collision evaluation;
\textbf{Scene}: evaluated scene representation.
LTL and Attr.\ refer to performance metrics,
not dataset statistics or annotations.
A cross indicates that the criterion is not reported
in the published protocol, not that occlusion is absent.
}
\end{table}

\begin{table*}[!t]
\centering
\caption{\textbf{Closed-loop results on OccEVT-Bench.}
SR and CFR are fractions; duration metrics are in seconds.}
\label{tab:main_benchmark}

\footnotesize
\setlength{\tabcolsep}{1pt}
\renewcommand{\arraystretch}{0.95}

\begin{tabular*}{\textwidth}{
@{\extracolsep{\fill}} l *{11}{r} @{}
}
\toprule
& \multicolumn{6}{c}{\textbf{Risk episodes}}
& \multicolumn{5}{c}{\textbf{Safe episodes}} \\
\cmidrule(lr){2-7}
\cmidrule(lr){8-12}

\textbf{Method}
& SR $\uparrow$
& VD $\downarrow$
& LTL $\downarrow$
& POL $\downarrow$
& OOF $\downarrow$
& CFR $\uparrow$
& SR $\uparrow$
& VD $\downarrow$
& LTL $\downarrow$
& OOF $\downarrow$
& CFR $\uparrow$ \\
\midrule

Human Teleoperation
& 0.925 & 3.324 & 1.213 & 1.221 & 0.988 & 0.950
& 0.980 & 0.132 & 0.113 & 0.116 & 0.980 \\

Adap-RPF$^\dagger$~\cite{situ2025adaprpf}
& 0.270 & 20.879 & 19.109 & 1.991 & 18.815 & 0.920
& 0.865 & 1.336 & 1.289 & 1.336 & 0.920 \\

Mono-RPF~\cite{ye2023robotpersonfollowing}
& 0.005
& 24.646
& 24.704
& 24.334
& \underline{2.419}
& \textbf{0.985}
& \textbf{0.950}
& 2.134
& \underline{1.128}
& \underline{1.136}
& \textbf{1.000} \\

OA-VAT~\cite{sun2026oavat}
& \underline{0.350}
& \underline{19.986}
& \underline{13.432}
& \underline{3.755}
& 16.041
& 0.690
& 0.730
& \textbf{0.542}
& \textbf{0.521}
& \textbf{0.521}
& \textbf{1.000} \\

MiniCPM-RobotTrack
& 0.020
& 21.735
& 17.388
& 17.536
& 8.337
& 0.880
& 0.870
& 1.980
& 1.653
& 1.920
& 0.960 \\

OmTrackVLA~\cite{omtrackvla2025}
& 0.010
& 22.311
& 19.862
& 13.152
& 9.119
& \underline{0.975}
& \underline{0.900}
& 1.740
& 1.397
& 1.720
& \underline{0.990} \\

\midrule
\textbf{VP-EVT (Ours)}
& \textbf{0.775}
& \textbf{4.235}
& \textbf{3.442}
& \textbf{1.715}
& \textbf{1.747}
& 0.905
& \underline{0.900}
& \underline{1.720}
& 1.405
& 1.725
& \underline{0.990} \\

\bottomrule
\end{tabular*}

\vspace{1pt}
\begin{minipage}{\textwidth}
\scriptsize
SR: success rate;
VD: visibility deficit;
LTL: longest target loss;
POL: physical-occlusion loss;
OOF: out-of-FOV loss;
CFR: collision-free rate.
$^\dagger$Adap-RPF additionally uses depth observations.
Human Teleoperation and Adap-RPF are reference methods
and are excluded from ranking.
\textbf{Bold} and \underline{underlined} values indicate
the best and second-best results among the remaining methods,
respectively; ties receive the same formatting.
\end{minipage}
\end{table*}

\subsection{Evaluation Metrics}
\label{sec:metrics}

\begingroup
\setlength{\abovedisplayskip}{5pt plus 1pt minus 1pt}
\setlength{\belowdisplayskip}{5pt plus 1pt minus 1pt}
\setlength{\abovedisplayshortskip}{3pt plus 1pt minus 1pt}
\setlength{\belowdisplayshortskip}{3pt plus 1pt minus 1pt}
\setlength{\jot}{2pt}

We assess both following outcomes and visibility throughout
execution. Success and collision rates characterize task
performance and safety, while visibility deficit, loss
duration, and loss attribution reveal whether the agent
reduces sustained occlusion rather than merely recovering
the target eventually.

Let $v_t\in[0,1]$ denote the rendered visible target fraction
relative to a target-only reference at the same viewpoint,
and let $f_t\in\{0,1\}$ indicate whether the target center
lies inside the camera field of view.
Here, $t$ indexes the $T$ recorded frames of an episode,
and $\Delta t$ is the simulation interval.
Duration-based metrics are computed per episode and
averaged within each evaluation split.

\textbf{Success Rate (SR).}
SR is the fraction of episodes that achieve stable following without collisions. Stable following requires sufficient target visibility, a robot--target distance within $[1.2,3.5]\,\mathrm{m}$, and a valid target-relative pose. Risk episodes require a stable following window after the predefined interaction, whereas Safe episodes require stable following during the final evaluation window. For $N$ episodes, $\mathrm{SR}=N^{-1}\sum_{i=1}^{N}s_i$, where $s_i\in\{0,1\}$ indicates episode success.

\textbf{Visibility Deficit (VD).}
VD measures accumulated visibility degradation as $\mathrm{VD}=\sum_{t=1}^{T}(1-v_t)\Delta t$. It captures both partial visibility reduction and severe target loss, with lower values indicating better visibility preservation.

\textbf{Longest Target Loss (LTL).}
Target loss occurs when visibility $v_t$ falls below the threshold $\tau_{\mathrm{loss}}$. We mark these frames with $\ell_t=1$ and set $\ell_t=0$ otherwise. LTL measures the longest continuous period of target loss within an episode and is zero if no loss occurs. Lower values indicate less persistent target loss.

\textbf{Loss Attribution.}
We partition thresholded target loss into physical-occlusion loss (POL) and out-of-FOV loss (OOF), computed as $\mathrm{POL}=\sum_{t=1}^{T}\ell_t f_t\Delta t$ and $\mathrm{OOF}=\sum_{t=1}^{T}\ell_t(1-f_t)\Delta t$, respectively. POL operationally records loss while the target center remains inside the field of view, whereas OOF records loss outside it. Reporting both reveals whether reduced in-view occlusion comes at the cost of out-of-view failures. LTL includes both types of target loss.

\textbf{Collision-Free Rate (CFR).}
CFR measures the fraction of episodes without collisions during execution, $\mathrm{CFR}=N^{-1}\sum_{i=1}^{N}(1-c_i)$, where $c_i\in\{0,1\}$ indicates whether any collision occurs in episode $i$. This metric assesses the safety of visibility-preserving motion alongside its tracking benefits.

\subsection{Expert Demonstrations}

Using the same platform, we construct
\textbf{OccEVT-Expert} from \textbf{20{,}000}
dynamic-occlusion training episodes.
Expert demonstrations and privileged annotations
provide supervision for occlusion anticipation
and visibility-preserving motion.
Training episodes are separate from the benchmark
evaluation set, and privileged information is
not required at deployment.

\begin{table*}[t]
\centering
\caption{\textbf{Ablation study on 200 Risk episodes.}
All variants predict and execute four controls per decision.
SR and CFR are fractions; duration metrics are in seconds.}
\label{tab:risk_ablation}

\footnotesize
\setlength{\tabcolsep}{1pt}
\renewcommand{\arraystretch}{0.95}

\begin{tabular*}{\textwidth}{
@{\extracolsep{\fill}} l *{5}{c} *{6}{r} @{}
}
\toprule
\textbf{Variant}
& Hist.
& Beliefs
& Prop.
& Plan
& Mem.
& SR $\uparrow$
& VD $\downarrow$
& LTL $\downarrow$
& POL $\downarrow$
& OOF $\downarrow$
& CFR $\uparrow$ \\
\midrule

Single-frame observation
& \ding{55} & \ding{51} & \ding{55}
& \ding{51} & \ding{51}
& 0.435
& 8.901
& 6.611
& 3.176
& 5.647
& 0.630 \\

w/o target--pedestrian beliefs
& \ding{51} & \ding{55} & --
& \ding{51} & \ding{51}
& 0.110
& 20.184
& 18.762
& 12.974
& 8.431
& 0.685 \\

w/o occlusion-time belief propagation
& \ding{51} & \ding{51} & \ding{55}
& \ding{51} & \ding{51}
& 0.475
& 5.200
& 3.988
& 2.725
& 2.421
& 0.545 \\

w/o following-waypoint conditioning
& \ding{51} & \ding{51} & \ding{51}
& \ding{55} & \ding{51}
& 0.425
& 4.547
& 4.198
& 3.567
& 1.929
& 0.485 \\

w/o recurrent maneuver state
& \ding{51} & \ding{51} & \ding{51}
& \ding{51} & \ding{55}
& 0.695
& 5.046
& 4.044
& 2.999
& 1.985
& 0.755 \\

\midrule

w/o predictive risk (reactive triggering)
& \ding{51} & \ding{51} & \ding{51}
& \ding{51} & \ding{51}
& 0.685
& 9.675
& 6.246
& 2.975
& 3.747
& 0.725 \\

w/o risk gating (always-on control)
& \ding{51} & \ding{51} & \ding{51}
& \ding{51} & \ding{51}
& 0.465
& 6.732
& 3.949
& 2.343
& 4.975
& 0.525 \\

\midrule

\textbf{Full VP-EVT}
& \ding{51} & \ding{51} & \ding{51}
& \ding{51} & \ding{51}
& \textbf{0.775}
& \textbf{4.235}
& \textbf{3.442}
& \textbf{1.715}
& \textbf{1.747}
& \textbf{0.905} \\

\bottomrule
\end{tabular*}

\vspace{1pt}
\begin{minipage}{\textwidth}
\scriptsize
Hist.: five-frame RGB history;
Beliefs: target--pedestrian beliefs;
Prop.: occlusion-time belief propagation;
Plan: following-waypoint conditioning;
Mem.: recurrent maneuver state.
\ding{51}/\ding{55} indicate retained/removed components;
-- indicates not applicable.
The two intervention variants retain all five listed components:
reactive triggering activates intervention after tracking
confidence deteriorates, whereas always-on control
continuously executes the intervention controller.
\textbf{Bold} indicates the best value in each column.
\end{minipage}
\end{table*}
% Required packages:
% \usepackage{booktabs,graphicx,pifont}

\section{EXPERIMENTS}
\label{sec:experiments}

We evaluate tracking, visibility preservation, and safety through benchmark comparisons, ablations, and a real-world demonstration.

\subsection{Experimental Setup}

\textbf{Benchmark and metrics.}
All methods are evaluated on 400 held-out OccEVT-Bench episodes with identical episode configurations, camera settings, robot dynamics, and termination rules. Risk and Safe results are reported separately. Benchmark construction and metrics are detailed in Sec.~\ref{sec:benchmark} and Sec.~\ref{sec:metrics}.

\textbf{Baselines.}
\label{sec:baseline}
We compare with Mono-RPF~\cite{ye2023robotpersonfollowing} for joint-based target localization, OA-VAT~\cite{sun2026oavat} for occlusion-aware tracking and recovery, MiniCPM-RobotTrack for image-and-language-conditioned velocity control, and OmTrackVLA~\cite{omtrackvla2025} for following-waypoint generation. OmTrackVLA also supplies VP-EVT's waypoints, enabling comparison with the base follower. Human teleoperation and depth-assisted Adap-RPF~\cite{situ2025adaprpf} serve as unranked references.

\textbf{Training and implementation.}
\label{sec:implement}
VP-EVT is trained on OccEVT-Expert with the visual encoder and base follower frozen. Training and evaluation episodes are disjoint, and privileged annotations are used only for supervision.

Habitat-GS simulations use a differential-drive robot with a $90^{\circ}$ horizontal-FOV camera. Simulation and control run at $10\,\mathrm{Hz}$. VP-EVT uses five RGB frames spanning $0.8\,\mathrm{s}$ and, during intervention, predicts and executes four controls over $0.4\,\mathrm{s}$ per decision.
% ============================================================

\subsection{Main Closed-Loop Results}
\label{sec:main_results}

\textbf{Overall tracking performance.}
As shown in Table~\ref{tab:main_benchmark}, VP-EVT
achieves a Risk SR of $77.5\%$, exceeding OA-VAT's
$35.0\%$ by $42.5$ percentage points.
VP-EVT builds on the following waypoints supplied
by OmTrackVLA, increasing Risk SR from $1.0\%$
to $77.5\%$ with the proposed visibility-preserving
components.
This improvement demonstrates the value of
augmenting an existing follower with explicit
occlusion handling rather than relying on its
following waypoints alone.

The contrast between Risk and Safe performance
further clarifies the challenge.
Mono-RPF and OmTrackVLA achieve Safe SRs of
$95.0\%$ and $90.0\%$, respectively, yet rarely
succeed on Risk episodes.
Their difficulty therefore emerges under the
occlusion-challenged interactions despite strong
ordinary-following performance.
VP-EVT substantially narrows this performance gap,
indicating that its benefits address a weakness
that ordinary-following evaluation alone would
largely conceal.

\textbf{Visibility preservation.}
The improvement in task completion is accompanied
by substantial reductions in target loss.
Relative to OA-VAT, VP-EVT reduces VD by $78.8\%$
and LTL by $74.4\%$, reaching $4.235\,\mathrm{s}$
and $3.442\,\mathrm{s}$, respectively.
These gains show that the higher SR is accompanied
by less accumulated visibility degradation and
shorter uninterrupted loss, rather than a change
in the final task outcome alone.

The loss decomposition reveals a further distinction.
OA-VAT records a POL of $3.755\,\mathrm{s}$,
but its OOF reaches $16.041\,\mathrm{s}$,
showing that substantial target loss remains
outside the camera field of view.
VP-EVT reduces these values to $1.715\,\mathrm{s}$
and $1.747\,\mathrm{s}$, respectively.
The largest absolute reduction is in OOF, while
in-view loss also decreases.
Its advantage therefore extends beyond reducing
recorded physical occlusion to limiting out-of-view
target loss.
The simultaneous improvements in POL, OOF, and
LTL provide complementary evidence that VP-EVT
improves visibility during tracking, rather than
favoring one loss category at the expense of another.

\textbf{Safety and standard following.}
On Safe episodes, VP-EVT matches OmTrackVLA
in SR ($90.0\%$) and CFR ($99.0\%$).
Their visibility metrics are also close:
VD is slightly lower for VP-EVT, while LTL and
OOF are slightly higher.
Thus, the substantial improvement on Risk episodes
is not accompanied by a comparable change in
ordinary-following performance.
This supports retaining the base follower's
aggregate performance when persistent occlusion
is absent, although it does not directly establish
how often unnecessary interventions occur.

The safety outcome on Risk episodes depends on
the comparison.
VP-EVT achieves a CFR of $90.5\%$, improving
on OA-VAT's $69.0\%$ while also achieving higher SR.
Relative to this baseline, better occlusion handling
is therefore accompanied by fewer collision episodes.
However, CFR remains below OmTrackVLA's $97.5\%$.
The base follower's high CFR, together with its
low Risk SR, highlights that collision-free
execution alone does not ensure successful tracking.
VP-EVT improves task completion and visibility
substantially, but preserving the base follower's
collision-free performance during these challenging
interactions remains an important limitation.

\begin{figure}[t]
    \centering
    \includegraphics[width=\linewidth]
    {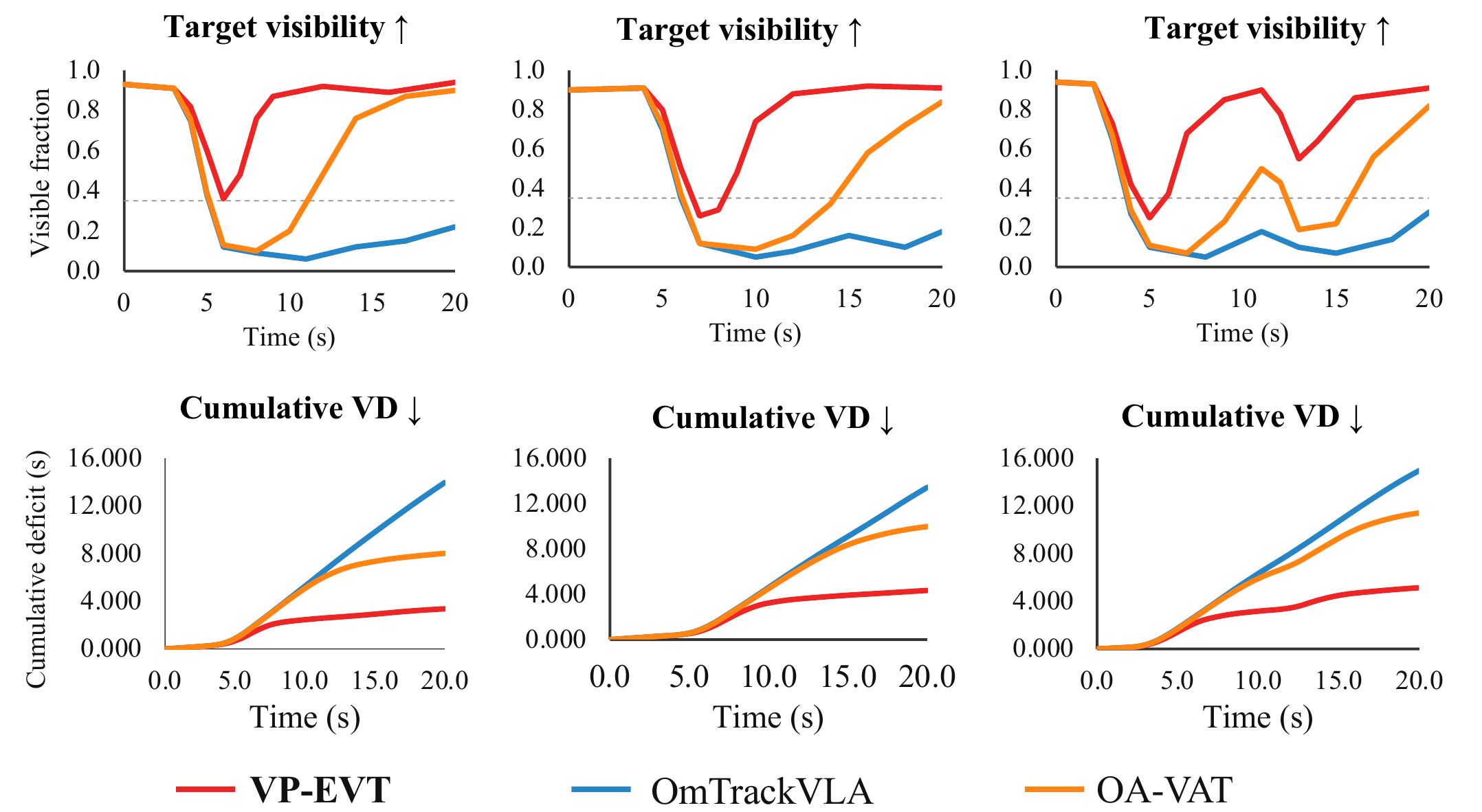}
    \caption{\textbf{Temporal visibility comparison.}
    We compare target visibility and cumulative visibility
    deficit across methods on selected Risk episodes.
    The dashed line marks the target-loss threshold
    $\tau_{\mathrm{loss}}=0.35$.
    Higher visibility and lower cumulative deficit
    indicate better visibility preservation.}
    \label{fig:temporal_visibility}
\end{figure}

\begin{figure}[t]
    \centering
    \includegraphics[width=\linewidth]{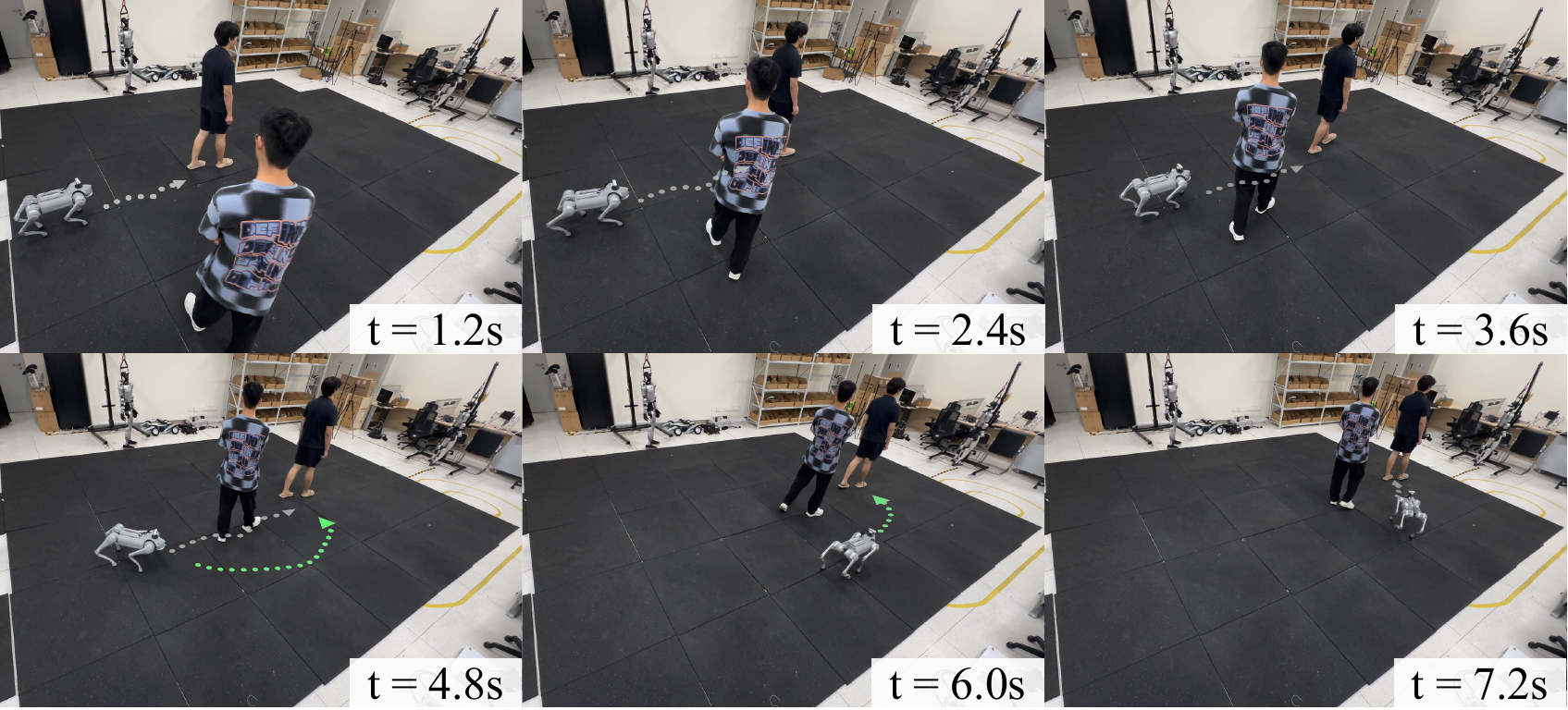}
    \caption{\textbf{Real-world deployment of VP-EVT.}
    The robot adjusts its viewpoint around the pedestrian
    in the patterned shirt to reduce occlusion and resume
    following the target in black.
    Gray dotted paths indicate the following reference,
    while green dotted paths illustrate visibility-preserving
    motion.
    Frames are ordered chronologically from left to right,
    top to bottom.}
    \label{fig:visualize}
    \vspace{-0.10in}
\end{figure}

% ============================================================
\subsection{Ablation Study}
\label{sec:ablation}

Table~\ref{tab:risk_ablation} evaluates the components
and intervention strategy of VP-EVT on 200 Risk episodes.
All variants follow the same evaluation protocol as
the main comparison and predict and execute four
controls per decision.

\textbf{Target--pedestrian beliefs.}
Removing both beliefs causes the largest decrease
in task success, reducing SR from 77.5\% to 11.0\%.
POL increases from $1.715$ to $12.974\,\mathrm{s}$,
while LTL rises from $3.442$ to $18.762\,\mathrm{s}$.
The deterioration in both task completion and
target-loss persistence supports explicit
target--pedestrian state representations for
visibility-preserving decisions.
This joint ablation establishes the importance
of the combined representation, rather than
isolating the contribution of either belief.

\textbf{Temporal evidence and belief propagation.}
Disabling occlusion-time belief propagation reduces
SR to 47.5\% and CFR to 54.5\%, compared with
77.5\% and 90.5\% for the full model.
These results support maintaining estimated states
through observation gaps.
The single-frame variant further reduces SR to 43.5\%,
with LTL reaching $6.611\,\mathrm{s}$ and OOF
$5.647\,\mathrm{s}$.
Since this variant also disables belief propagation,
the comparison reflects their joint removal and
does not isolate the effect of visual history alone.

\textbf{Predictive risk and selective intervention.}
Reactive triggering reduces SR from $77.5\%$ to $68.5\%$, while increasing VD from $4.235$ to $9.675\,\mathrm{s}$ and LTL from $3.442$ to $6.246\,\mathrm{s}$. Waiting for tracking confidence to deteriorate therefore incurs greater visibility loss despite eventual recovery.

Continuous intervention reduces SR to $46.5\%$ and CFR to $52.5\%$, while increasing OOF from $1.747$ to $4.975\,\mathrm{s}$. Together, these results support anticipating occlusion and intervening selectively, rather than reacting only after visibility degrades or replacing standard following throughout the episode.

\textbf{Waypoint conditioning and maneuver memory.}
Removing waypoint conditioning reduces SR to $42.5\%$ and CFR to $48.5\%$. Resetting maneuver memory lowers SR to $69.5\%$ and CFR to $75.5\%$ and worsens all four visibility-loss metrics, despite retaining visual history, belief propagation, and the four-control execution horizon. These results support conditioning intervention on the following reference and preserving controller context across decisions. Duration metrics should be interpreted alongside SR and CFR, since early collision termination shortens the recorded evaluation window.

\subsection{Real-World Deployment}
\label{sec:qualitative}

We deploy VP-EVT on a Unitree Go2 equipped with a RealSense D435i camera, with inference performed on an off-board computer featuring an Intel Core i9-13900K CPU and an NVIDIA GeForce RTX 3090 GPU. Figure~\ref{fig:visualize} illustrates a pedestrian obstructing the robot's view of the target during following. The robot moves laterally around the pedestrian, reduces the obstruction as it advances, and subsequently resumes following from a suitable position behind the target.

\section{CONCLUSION}

We presented \textbf{VP-EVT}, which treats target visibility as a motion-control objective rather than a by-product of following. Language-guided temporal beliefs support waypoint-conditioned occlusion prediction and selective recurrent control to preserve visibility and restore stable following. We introduced \textbf{OccEVT-Bench} and \textbf{OccEVT-Expert} for evaluation and training. Experiments show higher tracking success and reduced visibility degradation and continuous target loss, with comparable Safe-episode performance to the base follower. Future work will improve intervention safety and robustness in denser crowds and diverse real-world settings.

%\addtolength{\textheight}{-12cm}

\bibliographystyle{IEEEtran}
\balance
\bibliography{IEEEabrv, ref}

\begin{thebibliography}{10}
\providecommand{\url}[1]{#1}
\csname url@samestyle\endcsname
\providecommand{\newblock}{\relax}
\providecommand{\bibinfo}[2]{#2}
\providecommand{\BIBentrySTDinterwordspacing}{\spaceskip=0pt\relax}
\providecommand{\BIBentryALTinterwordstretchfactor}{4}
\providecommand{\BIBentryALTinterwordspacing}{\spaceskip=\fontdimen2\font plus
\BIBentryALTinterwordstretchfactor\fontdimen3\font minus \fontdimen4\font\relax}
\providecommand{\BIBforeignlanguage}[2]{{%
\expandafter\ifx\csname l@#1\endcsname\relax
\typeout{** WARNING: IEEEtran.bst: No hyphenation pattern has been}%
\typeout{** loaded for the language `#1'. Using the pattern for}%
\typeout{** the default language instead.}%
\else
\language=\csname l@#1\endcsname
\fi
#2}}
\providecommand{\BIBdecl}{\relax}
\BIBdecl

\bibitem{luo2018active}
W.~Luo, P.~Sun, F.~Zhong, W.~Liu, T.~Zhang, and Y.~Wang, ``End-to-end active object tracking via reinforcement learning,'' in \emph{Proc. Int. Conf. Mach. Learn.}, 2018, pp. 3286--3295.

\bibitem{zhong2021distraction}
F.~Zhong, P.~Sun, W.~Luo, T.~Yan, and Y.~Wang, ``Towards distraction-robust active visual tracking,'' in \emph{Proc. Int. Conf. Mach. Learn.}, 2021, pp. 12\,782--12\,792.

\bibitem{zhong2024evt}
F.~Zhong, K.~Wu, H.~Ci, C.~Wang, and H.~Chen, ``Empowering embodied visual tracking with visual foundation models and offline {RL},'' in \emph{Proc. Eur. Conf. Comput. Vis.}, 2024, pp. 139--155.

\bibitem{wang2025trackvla}
S.~Wang, J.~Zhang, M.~Li, J.~Liu, A.~Li, K.~Wu, F.~Zhong, J.~Yu, Z.~Zhang, and H.~Wang, ``{TrackVLA}: Embodied visual tracking in the wild,'' in \emph{Proc. Conf. Robot Learn.}, 2025, pp. 4139--4164.

\bibitem{maalouf2024followanything}
A.~Maalouf, N.~Jadhav, K.~M. Jatavallabhula, M.~Chahine, D.~M. Vogt, R.~J. Wood, A.~Torralba, and D.~Rus, ``Follow anything: Open-set detection, tracking, and following in real-time,'' \emph{{IEEE} Robot. Autom. Lett.}, vol.~9, no.~4, pp. 3283--3290, 2024.

\bibitem{ye2023robotpersonfollowing}
H.~Ye, J.~Zhao, Y.~Pan, W.~Chen, L.~He, and H.~Zhang, ``Robot person following under partial occlusion,'' in \emph{Proc. {IEEE} Int. Conf. Robot. Autom.}, 2023, pp. 7591--7597.

\bibitem{ye2024personreidentification}
H.~Ye, J.~Zhao, Y.~Zhan, W.~Chen, L.~He, and H.~Zhang, ``Person re-identification for robot person following with online continual learning,'' \emph{{IEEE} Robot. Autom. Lett.}, vol.~9, no.~11, pp. 9151--9158, 2024.

\bibitem{liu2025trackvlapp}
J.~Liu, Y.~Qi, J.~Zhang, M.~Li, S.~Wang, K.~Wu, H.~Ye, H.~Zhang, Z.~Chen, F.~Zhong, Z.~Zhang, and H.~Wang, ``{TrackVLA++}: Unleashing reasoning and memory capabilities in {VLA} models for embodied visual tracking,'' \emph{arXiv:2510.07134}, 2025.

\bibitem{ye2025rpfsearch}
H.~Ye, K.~Cai, Y.~Zhan, B.~Xia, A.~Ajoudani, and H.~Zhang, ``{RPF-Search}: Field-based search for robot person following in unknown dynamic environments,'' \emph{{IEEE/ASME} Trans. Mechatronics}, vol.~30, no.~6, pp. 4129--4141, 2025.

\bibitem{sun2026oavat}
H.~Sun, K.~Zhou, H.~Gao, S.~Zhang, J.~Hu, X.~Wen, Q.~Ye, and M.~Tan, ``Instance-level visual active tracking with occlusion-aware planning,'' in \emph{Proc. {IEEE/CVF} Conf. Comput. Vis. Pattern Recognit.}, 2026, pp. 42\,494--42\,504.

\bibitem{wu2025vlmassistant}
K.~Wu, S.~Xu, H.~Chen, C.~Wang, Z.~Li, Y.~Wang, and F.~Zhong, ``{VLM} can be a good assistant: Enhancing embodied visual tracking with self-improving vision-language models,'' \emph{arXiv:2505.20718}, 2025.

\bibitem{luo2020endtoend}
W.~Luo, P.~Sun, F.~Zhong, W.~Liu, T.~Zhang, and Y.~Wang, ``End-to-end active object tracking and its real-world deployment via reinforcement learning,'' \emph{{IEEE} Trans. Pattern Anal. Mach. Intell.}, vol.~42, no.~6, pp. 1317--1332, 2020.

\bibitem{dionigi2024dvat}
A.~Dionigi, S.~Felicioni, M.~Leomanni, and G.~Costante, ``{D-VAT}: End-to-end visual active tracking for micro aerial vehicles,'' \emph{{IEEE} Robot. Autom. Lett.}, 2024, doi: 10.1109/LRA.2024.3385700.

\bibitem{zhong2019advat}
\BIBentryALTinterwordspacing
F.~Zhong, P.~Sun, W.~Luo, T.~Yan, and Y.~Wang, ``{AD-VAT}: An asymmetric dueling mechanism for learning visual active tracking,'' in \emph{Proc. Int. Conf. Learn. Represent.}, 2019. [Online]. Available: \url{https://openreview.net/forum?id=HkgYmhR9KX}
\BIBentrySTDinterwordspacing

\bibitem{zhong2021advatplus}
------, ``{AD-VAT+}: An asymmetric dueling mechanism for learning and understanding visual active tracking,'' \emph{{IEEE} Trans. Pattern Anal. Mach. Intell.}, vol.~43, no.~5, pp. 1467--1482, 2021.

\bibitem{zhang2025uninavid}
J.~Zhang, K.~Wang, S.~Wang, M.~Li, H.~Liu, S.~Wei, Z.~Wang, Z.~Zhang, and H.~Wang, ``{Uni-NaVid}: A video-based vision-language-action model for unifying embodied navigation tasks,'' in \emph{Proc. Robot. Sci. Syst.}, 2025, doi: 10.15607/RSS.2025.XXI.013.

\bibitem{liu2026comatrack}
Y.~Liu, L.~Gao, L.~Liu, M.~Lv, and Y.~Cai, ``{CoMaTrack}: Competitive multi-agent game-theoretic tracking with vision-language-action models,'' \emph{arXiv:2603.22846}, 2026.

\bibitem{wu2026adatracker}
K.~Wu, H.~Chen, J.~Han, H.~Liu, C.~Wang, Y.~Wang, Z.~Li, S.~Liu, and F.~Zhong, ``{AdaTracker}: Learning adaptive in-context policy for cross-embodiment active visual tracking,'' \emph{{IEEE} Robot. Autom. Lett.}, 2026, doi: 10.1109/LRA.2026.3692075.

\bibitem{dionigi2022evat}
A.~Dionigi, A.~Devo, L.~Guiducci, and G.~Costante, ``{E-VAT}: An asymmetric end-to-end approach to visual active exploration and tracking,'' \emph{{IEEE} Robot. Autom. Lett.}, vol.~7, no.~2, pp. 4259--4266, 2022.

\bibitem{yang2023policy}
P.~Yang, S.~Koga, A.~Asgharivaskasi, and N.~Atanasov, ``Policy learning for active target tracking over continuous {$SE(3)$} trajectories,'' in \emph{Proc. Learn. Dyn. Control Conf.}, 2023, pp. 64--75.

\bibitem{zhong2023rspt}
F.~Zhong, X.~Bi, Y.~Zhang, W.~Zhang, and Y.~Wang, ``{RSPT}: Reconstruct surroundings and predict trajectory for generalizable active object tracking,'' in \emph{Proc. {AAAI} Conf. Artif. Intell.}, vol.~37, no.~3, 2023, pp. 3705--3714.

\bibitem{ji2022elastic}
\BIBentryALTinterwordspacing
J.~Ji, N.~Pan, C.~Xu, and F.~Gao, ``{Elastic Tracker}: A spatio-temporal trajectory planner for flexible aerial tracking,'' in \emph{Proc. {IEEE} Int. Conf. Robot. Autom.}, 2022. [Online]. Available: \url{https://arxiv.org/abs/2109.07111}
\BIBentrySTDinterwordspacing

\bibitem{situ2025adaprpf}
W.~Situ, H.~Ye, J.~Peng, Y.~Zhan, and H.~Zhang, ``{Adap-RPF}: Adaptive trajectory sampling for robot person following in dynamic crowded environments,'' \emph{arXiv:2510.11308}, 2025.

\bibitem{hu2026cstwm}
J.~Hu, S.~Yuan, and Y.~Fang, ``{CST-WM}: A causally structured world model for embodied visual tracking,'' \emph{arXiv:2609.06302}, 2026.

\bibitem{fan2019lasot}
\BIBentryALTinterwordspacing
H.~Fan, L.~Lin, F.~Yang, P.~Chu, G.~Deng, S.~Yu, H.~Bai, Y.~Xu, C.~Liao, and H.~Ling, ``{LaSOT}: A high-quality benchmark for large-scale single object tracking,'' in \emph{Proc. {IEEE/CVF} Conf. Comput. Vis. Pattern Recognit.}, 2019. [Online]. Available: \url{https://arxiv.org/abs/1809.07845}
\BIBentrySTDinterwordspacing

\bibitem{huang2021got10k}
L.~Huang, X.~Zhao, and K.~Huang, ``{GOT-10k}: A large high-diversity benchmark for generic object tracking in the wild,'' \emph{{IEEE} Trans. Pattern Anal. Mach. Intell.}, vol.~43, no.~5, pp. 1562--1577, 2021.

\bibitem{sahin2023hoot}
G.~Sahin and L.~Itti, ``{HOOT}: Heavy occlusions in object tracking benchmark,'' in \emph{Proc. {IEEE/CVF} Winter Conf. Appl. Comput. Vis.}, 2023.

\bibitem{vanhoorick2023tcow}
B.~Van~Hoorick, P.~Tokmakov, S.~Stent, J.~Li, and C.~Vondrick, ``Tracking through containers and occluders in the wild,'' in \emph{Proc. {IEEE/CVF} Conf. Comput. Vis. Pattern Recognit.}, 2023.

\bibitem{ye2025tptbench}
H.~Ye, Y.~Zhan, W.~Situ, G.~Chen, J.~Yu, Z.~Zhao, K.~Cai, A.~Ajoudani, and H.~Zhang, ``{TPT-Bench}: A large-scale, long-term and robot-egocentric dataset for benchmarking target person tracking,'' \emph{Int. J. Robot. Res.}, 2026, doi: 10.1177/02783649261447308.

\bibitem{zhong2025unrealzoo}
\BIBentryALTinterwordspacing
F.~Zhong, K.~Wu, C.~Wang, H.~Chen, H.~Ci, Z.~Li, and Y.~Wang, ``{UnrealZoo}: Enriching photo-realistic virtual worlds for embodied {AI},'' in \emph{Proc. {IEEE/CVF} Int. Conf. Comput. Vis.}, 2025. [Online]. Available: \url{https://github.com/UnrealZoo/unrealzoo-gym}
\BIBentrySTDinterwordspacing

\bibitem{ye2025followbench}
H.~Ye, W.~Situ, J.~Peng, Y.~Zhan, B.~Xia, K.~Cai, and H.~Zhang, ``{Follow-Bench}: A unified motion planning benchmark for socially-aware robot person following,'' \emph{arXiv:2509.10796}, 2025.

\bibitem{omtrackvla2025}
K.~Lee, H.~Ying, and T.~Zhao, ``{OmTrackVLA}: Open-source visual language action model for visual navigation and following,'' \url{https://github.com/om-ai-lab/OmTrackVLA}, 2025.

\bibitem{xia2026habitatgs}
Z.~Xia, J.~Xu, C.~Cui, Y.~Yu, J.~Zhang, Q.~Yan, T.~Ni, J.~Chen, X.~Zhou, H.~Bao, R.~Hu, and S.~Peng, ``Habitat-gs: A high-fidelity navigation simulator with dynamic gaussian splatting,'' arXiv:2604.12626, 2026.

\end{thebibliography}
\end{document}